\documentclass[acmsmall,authorversion]{acmart}
\usepackage{multirow}
\usepackage{color, colortbl}
\usepackage{subcaption}
\usepackage{graphicx}
\usepackage{csquotes}
\newcommand{\textgroup}[1]{\textcolor{green!30!black}{\textit{{#1}}}}
\newcommand{\texttrait}[1]{\textcolor{purple!70!black}{\texttt{{#1}}}}
\begin{document}

\title{Multilingual large language models leak human stereotypes across language boundaries}

\author{Yang Trista Cao}
\authornote{Both authors contributed equally to this research.}
\author{Anna Sotnikova}
\authornotemark[1]
\email{aasotniko@gmail.com}
\affiliation{%
  \institution{University of Maryland}
  \country{USA}
}

\author{Jieyu Zhao}
\affiliation{%
  \institution{University of Southern California}
  \city{the work was done while at the University of Maryland}
  \country{USA}
 }

\author{Linda X. Zou}
\affiliation{%
 \institution{University of Maryland}
 \country{USA}
 }

\author{Rachel Rudinger}
\affiliation{%
  \institution{University of Maryland}
  \country{USA}
  }

\author{Hal Daum\'e III}
\affiliation{%
  \institution{University of Maryland, Microsoft Research}
   \country{USA}
  }

\renewcommand{\shortauthors}{Cao and Sotnikova, et al.}

\begin{abstract}
  Multilingual large language models have gained prominence for their proficiency in processing and generating text across languages. Like their monolingual counterparts, multilingual models are likely to pick up on stereotypes and other social biases present in their training data. In this paper, we study a phenomenon we term ``stereotype leakage'', which refers to how training a model multilingually may lead to stereotypes expressed in one language showing up in the models' behavior in another. We propose a measurement framework for stereotype leakage and investigate its effect across English, Russian, Chinese, and Hindi and with GPT-3.5, mT5, and mBERT. Our findings show a noticeable leakage of positive, negative, and non-polar associations across all languages. We find that of these models, GPT-3.5 exhibits the most stereotype leakage, and Hindi is the most susceptible to leakage effects.\\
  \textbf{WARNING: This paper contains model outputs which could be offensive in nature.}

\end{abstract}





\maketitle

\section{Introduction}
Large language models (LLMs) are trained on existing language data, and monolingual language models have been demonstrated to replicate stereotypical associations present in the training data.~\citep{stereoset, crows, our_stereo_2022}. 
Multilingual large language models (MLLMs) are language models pre-trained with a large amount of data from multiple languages so that they can process NLP tasks in various languages as well as cross-lingual tasks.
Previously, many studies have examined Western stereotypes in English language models~\citep[e.g.][]{stereoset, crows, our_stereo_2022}, whereas limited work has attempted to assess stereotypes in multilingual language models~\citep[e.g.][]{kaneko-etal-2022-gender, levy2023comparing, câmara2022mapping} due to the complexity of stereotypes manifested in various cultures, limited resources, and Anglocentric norms \citep{talat2022you}. 
However, the study of stereotypes in multilingual settings is more complex than examining them within each language. Due to the shared knowledge between languages in MLLMs, it is likely that stereotypes expressed in one language are transmitted to the models’ behavior in another language.

In this paper, we investigate the existence of \textit{stereotype leakage} in MLLMs. We define \textit{stereotype leakage} as the effect of stereotypical word associations in MLLMs of one language impacted by stereotypes from other languages. We conduct a human study to collect human stereotypes, adopt word association measurement approaches from previous studies ~\citep{our_stereo_2022, ilps} to measure stereotypical associations in MLLMs and analyze the strength and nature of stereotype leakage across different languages both quantitatively and qualitatively. 

Recent advancements in MLLMs have made them increasingly language-agnostic. For instance, models from GPT-family and mBART \citep{lin-etal-2022-shot} can operate without being restricted to a specific language, simultaneously handling input and output in multiple languages. This thus gives rising opportunities for what we refer to as stereotype leakage from one culture to another.\footnote{Although language models are trained on language-based data rather than culture-based data, languages inherently reflect the stereotypes associated with their respective cultures. To study stereotypes in MLLMs, we divide the world by languages, recognizing that a single language may represent multiple cultures.} Cultural stereotypes about social groups are shaped based on how these social groups are represented, treated, and discussed within each culture \citep{Martinez_2021, Lamer_2022, Rhodes_2012}. 
Hence, people's stereotypes about groups can be impacted by exposure to products and ideas from outside their own cultures. MLLMs, being the backbone of many natural language processing (NLP) applications, have the potential to exacerbate this issue by exporting harmful stereotypes across cultures and reinforcing Anglocentrism \cite{talat2022you,joshi-etal-2020-state}.\footnote{Anglocentrism is the practice of viewing and interpreting the world from an English-speaking perspective with the prioritization of English culture, language, and values. Anglocentrism can lead to biases and neglect of global perspectives and experiences.} 

We investigate the degree of stereotype leakage in MLLMs as a step toward understanding and mitigating this issue in AI systems. We test our hypothesis of significant stereotype leakage across languages in MLLMs by sampling four languages: English, Russian, Chinese, and Hindi. We choose languages from different writing systems---Latin alphabet, Cyrillic alphabet, Chinese characters, and Devanagari script---to enable a comprehensive evaluation of stereotype leakages in MLLMs. The models we assess are mBERT, mT5, and GPT-3.5.
Based on our findings, all models demonstrate varying degrees of stereotype leakage, which occurs bidirectionally across languages without a dominant directionality. Among the models tested, GPT-3.5 exhibits the highest degree of stereotype leakage. Importantly, the stereotype leakage includes not only negative stereotypes but also positive and non-polar associations. Furthermore, our study reveals that stereotypes in other languages about social groups unfamiliar to those cultures are shaped by the stereotypes present in the native language. This indicates that multilingual language models reflect and propagate cultural stereotypes across linguistic boundaries.

\begin{figure}
\centering
\includegraphics[width=\textwidth]{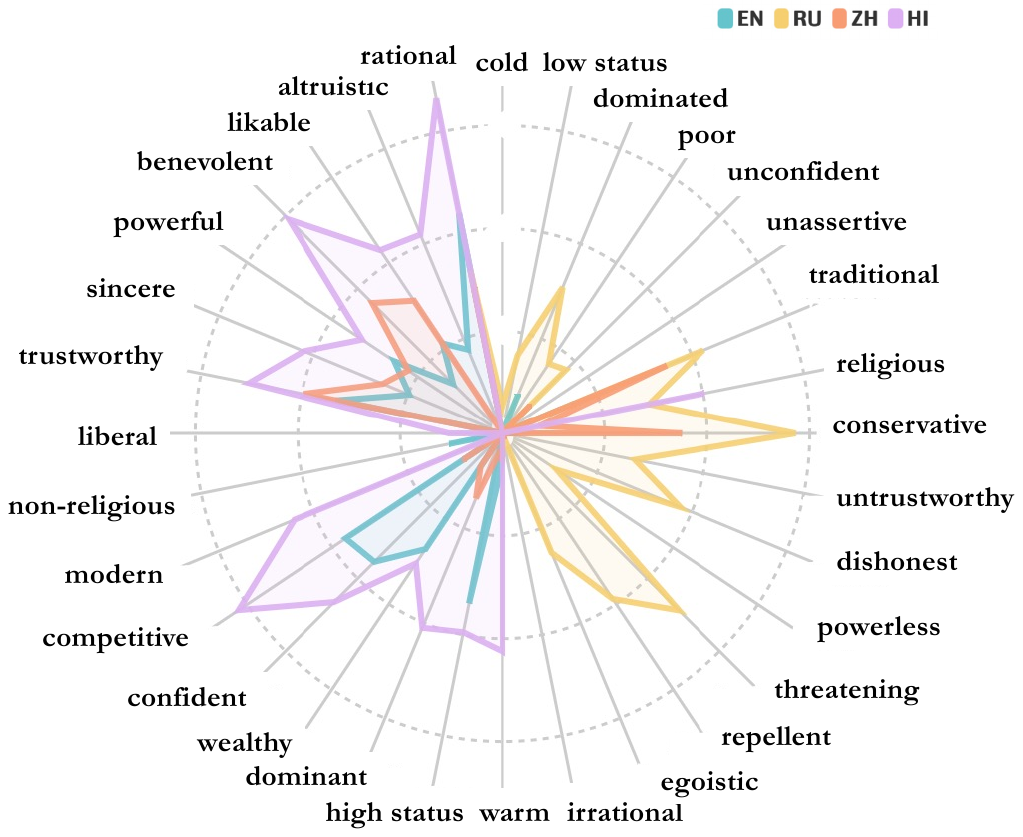}
\caption{The figure shows results of human annotations in English (EN), Russian (RU), Chinese (ZH), and Hindi (HI) languages based on the ABC model for the social group \textgroup{Asian people}. It shows average scores across all annotators per language.}
\label{fig:radar}
\end{figure}

\begin{table*}
    \centering
    \footnotesize
    \begin{tabular}{cr@{~$\leftrightarrow$~}ll|cr@{~$\leftrightarrow$~}l|lcr@{~$\leftrightarrow$~}l}
    \hline
    \multirow{6}{*}{\rotatebox{90}{\textbf{Agency}}} &
    powerless & powerful &
    &\multirow{6}{*}{\rotatebox{90}{\textbf{Beliefs}}} &
    \multicolumn{2}{c|}{} &
    &\multirow{6}{*}{\rotatebox{90}{\textbf{Communion}}} &
    untrustworthy & trustworthy \\
    &low status & high status &
    &&religious & non-religious  &
    &&dishonest & sincere \\
    &dominated & dominating &
    &&irrational & rational &
    &&cold & warm \\
    &poor & wealthy &
    &&conservative & liberal &
    &&threatening & benevolent\\
    &unconfident & confident & 
    &&traditional & modern &
    &&repellent & likable \\
    &unassertive & competitive &
    &&\multicolumn{2}{c|}{} &
    &&egotistic & altruistic \\
    \hline
    \end{tabular}
\caption{List of stereotype dimensions and corresponding traits in the ABC model; figure from \citep{our_stereo_2022}. 
}\label{tab:koch-traits}
\end{table*}

\section{Background and Related Work}

Assessing multi-cultural biases and stereotypes in multilingual settings is challenging. As noted by \citet{talat2022you}, there is a significant lack of benchmark datasets for measuring multilingual fairness. While many datasets exist in English, simply translating these datasets poses issues due to linguistic and cultural disparities. Furthermore, many existing fairness evaluation datasets are rooted in Western cultures, resulting in a gap that fails to encompass global cultural perspectives. \citet{bartl-etal-2020-unmasking} also highlighted the difficulty of measuring gender biases in languages with rich morphology and gender marking.

Many studies thus have been devoted to expanding the language boundary to assess the presence and impact of biases in multilingual settings by proposing new measurement approaches and evaluation datasets. \citet{wang2022} focused on evaluating the multilingual fairness of pre-trained multimodal representations. Many studies delve deeply into gender biases in multilingual settings. \citet{zhao-etal-2020-gender} focused on word representations, while both \citet{kaneko-etal-2022-gender} and \citet{steinborn-etal-2022-information} investigated gender bias in masked language models, each proposing new datasets for analyses. Furthermore, \citet{touileb-etal-2022-occupational} examined occupational biases within Norwegian and multilingual language models, seeking to identify and mitigate these biases. Addressing intersectional biases, \citet{câmara2022mapping} mapped biases in sentiment analysis systems across English, Spanish, and Arabic, proposing a framework to measure these biases effectively. Additionally, \citet{neveol-etal-2022-french} extended the CrowS dataset \cite{crows} of sentence pairs in English for measuring bias in masked language models to the French language.

Going further, \citet{dev-etal-2023-building} emphasized cultural inclusiveness by developing a stereotype dataset centered on Indian culture. This work highlights the importance of capturing local cultural contexts through community engagement. On the other hand, \citet{levy2023comparing} compared biases arising from multilingual training. Their findings show that biases are influenced by cultural contexts and often amplified during multilingual fine-tuning, underscoring the complexities involved in achieving fairness in multilingual NLP systems.

\section{Measuring Stereotype Leakage in MLLMs}

In measuring stereotype leakage in MLLMs, we aim to evaluate the extent to which stereotypes present in one language (the \emph{source language} of stereotype leakage) become manifested in the models' behavior in another language (the \emph{target language} of stereotype leakage) as a consequence of multilingual training. Specifically, we measure the effect of stereotypical word associations in MLLMs of the target language impacted by stereotypes from source languages. These stereotypical associations reflect stereotypes manifested in the models' behavior. In multilingual training, these can originate from stereotypes of the target language itself or from other languages. Meanwhile, some of these stereotypes are also captured during monolingual training. In this study, we aim to measure the effect of stereotype leakage from other languages as a result of multilingual training.

To do so, we form \autoref{eqn:equation}, where $\textit{MLLM}_{\textrm{tgt}}$ is the stereotypical word associations in MLLMs of the target language and $\textit{H}_{\textrm{en}}, \textit{H}_{\textrm{ru}}, \textit{H}_{\textrm{zh}}, \textit{H}_{\textrm{hi}}$ are human stereotypes of source languages. We measure the effect of stereotype leakage from source languages ($\textit{H}_{*}$) to the target language of the MLLMs ($\textit{MLLM}_{\textrm{tgt}}$). In this study, we sample four languages to study stereotype leakage: English (EN), Russian (RU), Chinese (ZH), and Hindi (HI). We choose languages that do not have shared orthography to focus on non-trivial leakages. These four languages come from the Indo-European and Sino-Tibetan language families, ranging from high (English) to low-resource (Hindi) languages. 

\begin{equation}
\begin{aligned}
\textit{MLLM}_{\textrm{tgt}} =& \alpha_{\textrm{en}}\textit{H}_{\textrm{en}} + \alpha_{\textrm{ru}}\textit{H}_{\textrm{ru}} + \alpha_{\textrm{zh}}\textit{H}_{\textrm{zh}}\\ 
& + \alpha_{\textrm{hi}}\textit{H}_{\textrm{hi}} + \beta \textit{LM}_{\textrm{tgt}}+ C \label{eqn:equation}
\end{aligned}
\end{equation}

$\textit{MLLM}_{\textrm{tgt}}$ and $\textit{H}_*$ are all $32\times 30$ dimensional matrices, where $32$ is the number of traits and $30$ is the number of social groups. Each entry in the matrices is the stereotypical association score for a specific trait of a social group. $C$ is the intercept. 

To measure the effect of stereotypes captured only due to multilingual training, we separate the stereotypes that may be captured during monolingual training by adding the $\textit{LM}_{\textrm{tgt}}$ variable, which is the stereotypical association from the target language's monolingual model. Similarly, $\textit{LM}_{\textrm{tgt}}$ is also a $32\times 30$ dimensional matrix. Note that since we can only find the monolingual BERT model for all four languages, we use these as proxies of $\textit{LM}_{\textrm{tgt}}$ for all MLLMs.\footnote{The monolingual BERT models we use are \href{https://huggingface.co/google-bert/bert-base-cased}{bert-
base-cased}, \href{https://huggingface.co/google-bert/bert-base-chinese}{bert-base-chinese}, \href{https://huggingface.co/cointegrated/rubert-tiny2}{cointegrated/rubert-tiny2}, and neuralspace-
reverie/indic-transformers-hi-bert.}

With the formula, the goal is to estimate how $\textit{H}$ of each language affects the $\textit{MLLM}_{\textrm{tgt}}$. We use a mixed-effect model to fit the formula and calculate the effect. The mixed effect model fits a linear regression to the formula with the data. In our case, we set traits as the random effect variable. With the fitted linear model, we get the coefficients ($\alpha$ and $\beta$). The $\alpha_*$ value indicates how much the $\textit{MLLM}_{\textrm{tgt}}$ stereotype score would change when the $\textit{H}_*$ score changes. If the coefficient ($\alpha$ and $\beta$) of a variable is positive and has a p-value of less than $0.05$, then we know that the variable has a significant effect on $\textit{MLLM}_{\textrm{tgt}}$. If there are significant effects from a non-target language's human stereotypes to $\textit{MLLM}_{\textrm{tgt}}$, then there is potential stereotype leakage from this non-target language to the target language.

\begin{table}
\centering\footnotesize
\begin{tabular}{m{1.4cm}>{\raggedright\arraybackslash}m{5.5cm}}
\hline
\textbf{Category} & \textbf{Groups} \\
\hline
\mbox{Shared/} \mbox{Shared}& man,  woman, gay,  lesbian, single mother, housewife, software engineer, wealthy person, poor person, disabled person\\[2em]
\mbox{Shared/} \mbox{Non-shared} & Asian person, Black person, Muslim person, immigrant, government official, civil servant, feminist, veteran\\[2em]
\mbox{Non-shared/} \mbox{Non-shared} & \textit{USA}: Texan,  Mormon,  Puerto Rican \mbox{\textit{Russia}: VDV soldier,  Muscovite, Chechenets} \mbox{\textit{China}: migrant worker, Hui person,} \mbox{Shanghainese person} \mbox{\textit{India}: Brahmin person, Gujarati person,} \mbox{Shudra person} 
\end{tabular}
\caption{Categories and corresponding social groups were used for the model and human experiments. ``Shared/Shared'' represents shared groups and shared stereotypes. ``Shared/Non-shared'' represents shared groups and non-shared stereotypes. ``Non-shared/Non-shared'' represents non-shared groups and non-shared stereotypes. 
}\label{tab:social_groups}
\end{table}

In our experiments, we focus on mBERT, mT5, and GPT-3.5.\footnote{The multilingual large language model versions we use are bert-base-multilingual-cased~\citep{mbert}, google/mt5-base~\citep{mt5}, and gpt-3.5-turbo-0125 version~\citep{ouyang2022training}.}
Both mBERT and mT5 are back-end MLLMs. MT5 has better multilingual performance than mBERT, whereas mBERT has more comparable monolingual BERT models for the four languages. GPT-3.5 is one of the state-of-the-art MLLMs that has been popularly deployed to users.

With these, we examine the effect of stereotype leakages in MLLMs.\footnote{The code and the dataset, along with a datasheet~\citep{datasheet}, will be published on GitHub.} In the following section, we discuss how we measure each of the variables.

\subsection{Stereotype Measurement}
In this paper, we adopt a theory-grounded stereotype measurement approach from \citet{our_stereo_2022}. We measure stereotypes through group-trait associations with traits from the Agency Beliefs Communion (ABC) model of stereotype content \citep{koch_2020}, a well-established stereotype
model constructed through social psychology studies for assessing human
stereotypes. The model consists of 16 trait pairs (each pair represents two polarities) that are designed to characterize group stereotypes along the dimensions of agency/socioeconomic success, conservative–progressive beliefs, and communion, as listed in \autoref{tab:koch-traits}. These trait pairs were selected to capture a broad range of stereotype dimensions to distinguish groups and are supported and validated by extensive literature in social psychology \citep{Koch_Yzerbyt_Abele_Ellemers_Fiske_2021,koch_comparison}.

If a group (e.g. \textgroup{immigrant, Asian person}) has a high degree of association with a trait (e.g. \texttrait{religious}, \texttrait{confident}), then we consider that trait a stereotype of the group. For example, \autoref{fig:radar} is the stereotype map of the group \textgroup{Asian people} collected from our human study across the four languages that we study.

For the groups, we picked $30$ groups listed in \autoref{tab:social_groups}: $10$ \textit{shared groups with shared stereotypes} (groups that are present in all four countries and are expected to be targeted by similar stereotypes), $8$ \textit{shared groups with non-shared stereotypes} (groups that are present in all four countries but expected to be targeted by dissimilar stereotypes), and $12$ \textit{non-shared groups} (groups that exist uniquely in each country). For shared groups, we manually selected groups from the list of social groups from \citet{our_stereo_2022} and categorized them as \textit{Shared/Shared} or \textit{Shared/Non-shared}. We verify the division with the human study. To collect non-shared groups, we surveyed native speakers. For each language, we asked $6$ native speakers to list $5-10$ social groups that they believe are unique to their culture. We then chose $3$ social groups per language based on the outcome of the majority vote.

In our human study, we further verify that each group matches the property of its category. To illustrate, stereotypes of groups in the first category exhibit an average correlation score of $0.60$ across languages. In contrast, groups in the second and third categories demonstrate progressively lower correlation scores of $0.50$ and $0.26$, respectively.

\subsection{Human stereotypes}
 \textbf{Survey Design:} To collect human stereotypes, we conducted a human study on Prolific\footnote{\url{https://www.prolific.co/}} with native speakers from the United States, Russia, China, and India\footnote{Approved by our institutional IRB.}. Participants were selected based on their status as either current or former residents of these countries and fluency in their respective native languages. The survey was administered in English for the U.S. and translated by native speakers into Chinese, Hindi, and Russian for the other regions. In the survey, participants first selected at least four social groups with which they were familiar. For each group, they were asked to rate associations between 16 trait pairs and the chosen social groups. We ensured a minimum of five annotations per social group in each language for both commonly shared social groups and shared groups that have unique stereotypes. For non-shared groups with unique stereotypes in a given language, we collected annotations only in that language. Each participant was compensated \$2.00 for evaluating five social groups on 16 trait pairs, which took on average 10 minutes to complete. For more details about the survey design, see Appendix \ref{sec:design}

\textbf{Annotation quality control:} 
Annotation quality verification for subjective tasks is a challenge as there is no ground truth exists. To ensure the maximum quality of the annotators, we implemented quality testing procedures. First, we selected only participants with a higher approval rate of 90\%, which we had to balance with the participants' availability. Then we added 3 test questions to the survey to verify their engagement and consistency. For more details about questions, see Appendix \ref{sec:quality}. A total of 286 participants completed the study, of which 151 passed the quality checks, emphasizing the importance of rigorous quality control procedures.

\textbf{Participants demographics:}
We collected demographic data on gender, age, education, and, for non-English speakers, their consumption of American social media, with participants free to skip questions. Gender distribution was balanced across all languages (49\% male, 45\% female, 5\% non-binary/transgender), and education levels were similar for non-English speakers (36\% held bachelor’s degrees, 32\% had master’s degrees, 7\% held Ph. D.s). English speakers had no Ph.D. holders and a higher proportion of high school graduates (35\%). Most participants were younger, with 42\% aged 18-30. Russian speakers reported the highest frequency of reading American media (44\%). More details on participants' demographics are available in Appendix \autoref{sec:demo}.

\subsection{Model stereotypical associations}
To measure stereotypical group-trait associations in large language models, we adopt different approaches for different MLLMs. For mT5, we use the increased log probability score (ILPS) \citep{ilps}, which computes the likelihood for the model to generate a trait given a templated sentence about a group. For example, $p(\texttrait{competent}~|~\textit{\textgroup{Asian people} are \rule{0.3cm}{0.15mm}.''})$ indicates the correlation between \texttrait{competent} and \textgroup{Asian people}.

For mBERT, we use the sensitivity test (SeT) \citep{our_stereo_2022}, which is shown to have better alignment with human stereotypes~\citep{our_stereo_2022}. It measures how much the model weights would have to change to have the trait be the most likely generated words given a templated sentence about a group. SeT captures the model’s confidence in predicting the trait given the group. 

\begin{figure*}[t]
    \centering
    \footnotesize
    \begin{tabular}{c@{\hskip1.8pt}c@{\hskip1.8pt}c@{\hskip1.8pt}c@{\hskip1.8pt}}
    & {\bfseries mBERT}& {\bfseries mT5}& {\bfseries GPT-3.5}\\
    \rotatebox[origin=c]{90}{\textbf{Flow Intensity}} &
    \raisebox{-0.5\height}{
        \begin{subfigure}{0.3\textwidth}
\includegraphics[width=0.95\textwidth,height=4.0cm]{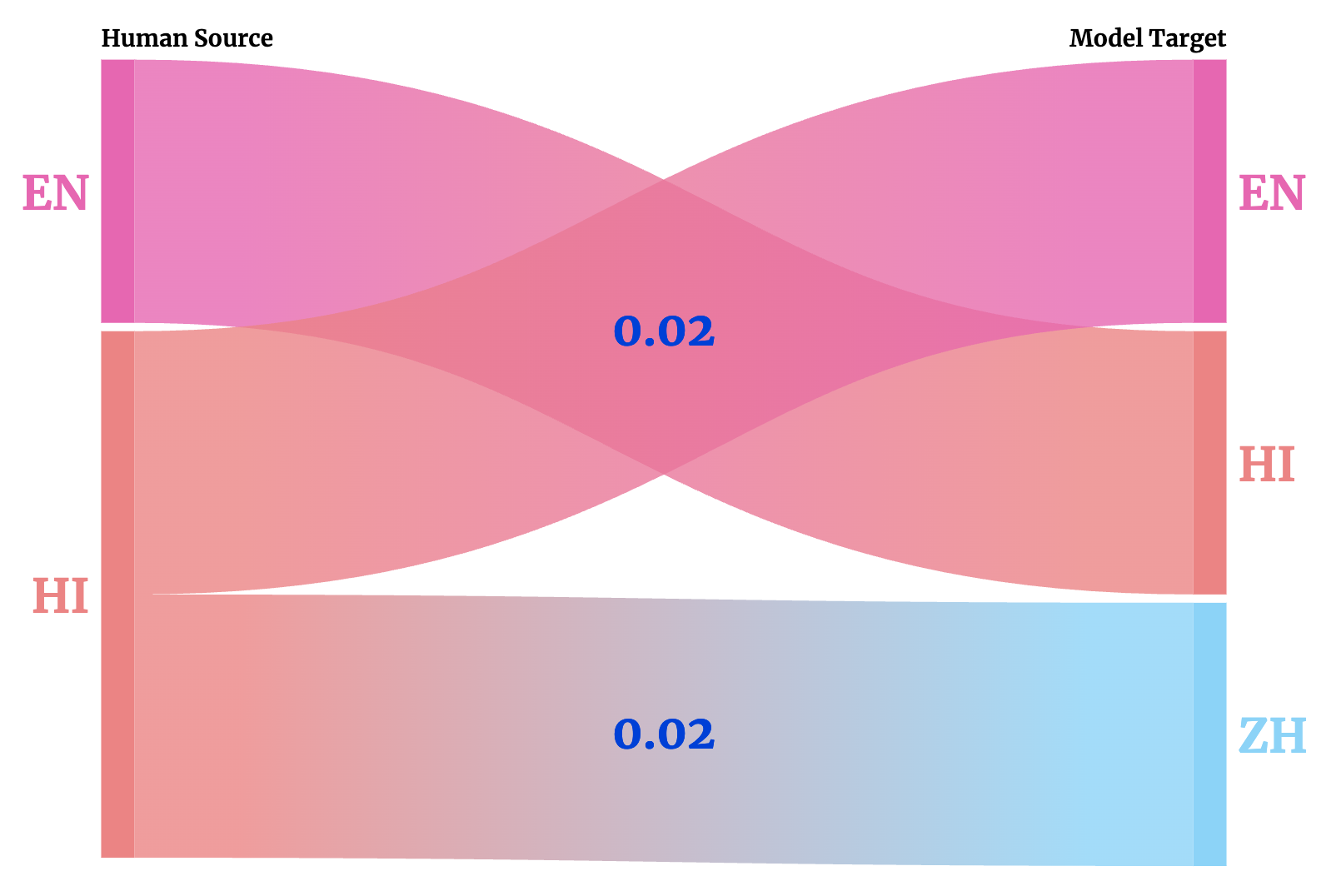}
        \end{subfigure}}&
    \raisebox{-0.5\height}{
        \begin{subfigure}{0.3\textwidth}
\includegraphics[width=0.95\textwidth,height=4.0cm]{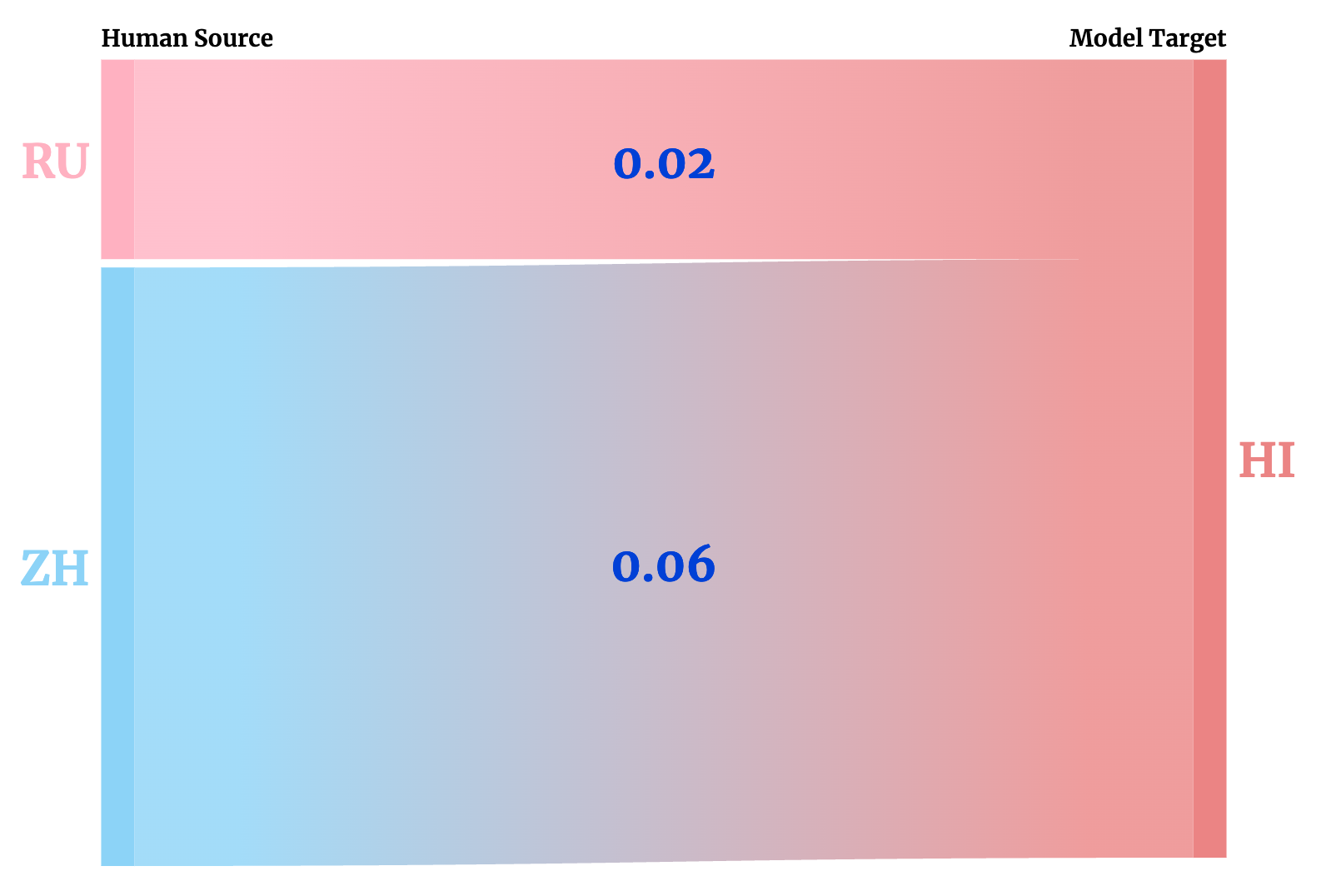}
        \end{subfigure}
    }&
    \raisebox{-0.5\height}{
        \begin{subfigure}{0.3\textwidth}
\includegraphics[width=0.95\textwidth,height=4.0cm]{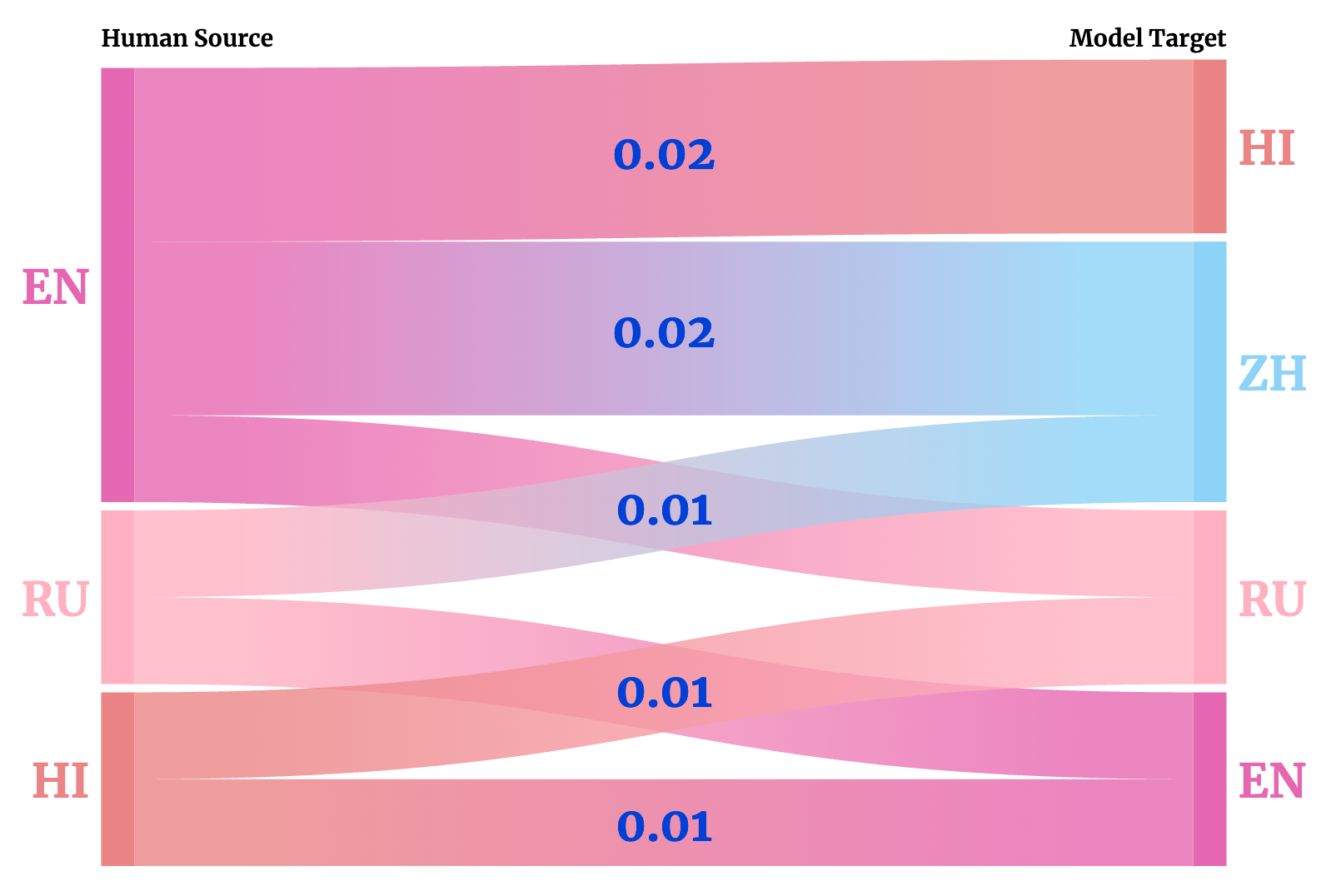}
        \end{subfigure}
    }
    \end{tabular}
    \caption{The figures show stereotype leakages for three models: mBERT, mT5, and GPT-3.5 respectively. Each figure illustrates the flow from the human source language (the left column) to the target language in a particular model (the right column). The numbers are the mixed-effect coefficients (denoted as $\alpha$ in \autoref{eqn:equation}). If no flow for a particular language is presented, this means that no significant leakage is happening.}\label{fig:side4} 
\end{figure*}

In evaluating GPT-3.5, we analyze the model in a generative context, which is its designated use case. We measure each group's association with various traits through pairwise comparisons. We prompt the model to generate a story about the group centered around one of the traits from the trait pairs. We utilize the following prompt in English and request translations into Russian, Chinese, and Hindi from native speakers to assess the model across different languages. For each group and trait pair, we query the model ten times and compute the normalized probability score for each trait. Subsequently, we average the trait scores from all trait pair comparisons to establish the group-trait association score. To ensure equitable comparisons among positive and negative traits, we juxtapose positive traits with positive ones and negative traits with negative ones.

\begin{description}
  \item[system prompt:] \texttt{You are a writer.}
  \item[prompt:] \texttt{You are writing about \textgroup{\{GROUP\}}. Before writing, think about what theme you want to pick. You can choose either `\texttrait{\{TRAIT1\}}' or `\texttrait{\{TRAIT2\}}' as your theme. You can also choose `neither' if you think neither of these themes fits. Note that you can choose only one theme. Output the exact name of the theme only, without any punctuation.}
\end{description}

\noindent
When processing the outputs of GPT-3.5, we employ an exact match criterion to assign scores to traits. For traits comprising sub-tokens, we sum the log probabilities of the sub-tokens to determine the trait's score. In instances where outputs do not precisely match the traits in the prompt, we have them manually processed by native speakers of the respective language. 

Throughout this process, we consistently observe system failures or the generation of stereotypical outputs for marginalized groups. Please refer to Appendix \ref{sec:failure} for some examples.  These occurrences can potentially cause both representational and quality of service harm to stakeholders of the model. While we do not explicitly analyze these patterns, we believe future research should investigate them thoroughly.

\begin{table}
  \centering
  \begin{tabular}{lcccc}
    \hline
     Mono-BERT  $\longrightarrow$ & \textbf{EN} & \textbf{RU} & \textbf{ZH} & \textbf{HI} \\
     \hline
     mBERT & 0.33 & 0.29 & 0.17 & 0.08 \\
    mT5 &0.10	&0.45	&0.14	&0.14 \\
    GPT-3.5 &0.07	&0.05	&0.05	&0.06\\
    \hline
  \end{tabular}
  \caption{Mixed-effect coefficients of monolingual BERTs (denoted as $\beta$ in \autoref{eqn:equation}) in the respective languages contributing to the same languages in multilingual language models. All of the effects are statistically significant. Note that the coefficients are not comparable across multilingual language models as the score ranges are different. }
  \label{tab:monobert}
\end{table}

\section{Stereotype Leakage and Its Effects}

In this section, we present our quantitative and qualitative results of the assessment of stereotype leakage across languages in MLLMs. 

Our quantitative analysis measures stereotype leakage by assessing how stereotypical associations in target language models are influenced by human stereotypes from source languages. We use mixed-effect models to quantify this leakage and identify significant cross-language influences. 

Qualitatively, we examine specific stereotypical associations that leak from one language to another, exploring their nature and implications. We investigate both positive and negative stereotypes\footnote{When we say ``positive stereotypes'', we mean stereotypical associations with positive traits. Note that stereotypes, regardless of how harmless they may seem, foster the essentialization of people}, as well as non-polar associations, to provide a comprehensive understanding of how stereotypes are transmitted across languages.

\subsection{Quantitative Results}

We compute the stereotype leakage across languages within three MLLMs based on \autoref{eqn:equation}.
The findings are presented in \autoref{fig:side4}, illustrating the extent to which stereotypical associations in the target language model are influenced by human stereotypes present in the culture associated with the source language. For example, in \autoref{fig:side4}, we observe that within GPT-3.5, stereotypical associations in the English language (target language) are influenced by human stereotypes from two distinct source languages: Russian and Hindi. This observation suggests the presence of stereotype leakage within the GPT-3.5 model.

In our analysis of mBERT, we observe significant leakages of stereotypes from Hindi to English and Chinese with coefficients of $0.02$ ($p=0.009$) and $0.06$ ($p=0.00$), respectively. We also observe English human stereotypes manifesting in mBERT Hindi with a coefficient of $0.02$ ($p=0.048$). Within the mT5 model, we find two significant stereotype leakages, both of which are leakages targeting Hindi. Russian and Chinese human stereotypes manifest in mT5 Hindi with coefficients of $0.02$ ($p=0.047$) and $0.06$ ($p=0.00$), respectively. For GPT-3.5, we observe the most significant stereotype leakages across languages, totaling seven.
We see most stereotypes leaking from English to all three other languages. The largest flows are from English to Chinese and Hindi, with coefficients of $0.02$ ($p=0.00$). Meanwhile, all languages are prone to be affected by leakages from other languages, even English.

Moreover, among all languages, Hindi experiences the highest degree of stereotype leakage --- it has four cases of significant stereotype leakage from other languages across three MLLMs. Since Hindi is the only low-resource language we tested, this might explain why it absorbs stereotypes from other languages. Both Chinese and English languages have three leakages across the models. The Chinese language has leakages from all three other languages, while the English language has the most leakage from Hindi. The Russian language has two significant leakages from English and Hindi.

Finally, we report the coefficients of effects from monolingual language models ($LM_{tgt}$) in \autoref{tab:monobert}. All the effects are statistically significant and are stronger than the effects from human stereotypes. This is not surprising because monolingual language models and multilingual language models share similar training data and model structures.

\subsection{Qualitative Results}

We then delve into the specific stereotypical associations that transfer from one language to another, considering the potential impact of such strengthened associations. Our focus is on the GPT-3.5 model, where we observe the highest degree of stereotype leakage. For each source-target language pair that exhibits significant stereotype leakage based on our results, we examine the traits most strongly associated with each group in the target language. We pay particular attention to traits that are not deemed associated to the group according to target language human stereotypes but align with stereotypes from the source language. Our analysis identifies two primary types of leakage: the amplification of positive and negative associations for certain languages. Additionally, we observe non-polar leakages, which are characterized by associations that are neither positive nor negative.

\subsubsection{Positive Leakage}
According to human annotation, \textgroup{Asian people} are more positively perceived in the English language than in Russian. We observe the strengthening of such traits in GPT-3.5 Russian language as \texttrait{wealthy, likable}, and \texttrait{high status}, possibly resulting from leakages from English and other languages. Moreover, \textgroup{housewives} become more \texttrait{warm} in English following leakages from possibly Russian and Hindi. \textgroup{Black people} are more \texttrait{powerful, modern, confident}, and \texttrait{wealthy} in the English language following leakage from Hindi. Another example of the leakage of positive perceptions is for \textgroup{gay men} and \textgroup{lesbians} from English to other languages. Traits such as \texttrait{likable, confident, warm, dominant, sincere}, and \texttrait{powerful} become stronger in Russian, Chinese, and Hindi. 

\subsubsection{Negative Leakage}
Meanwhile, there are negative stereotypes that leak across languages. From \textgroup{feminists}, we observe a leakage from English to Chinese and Hindi, and from Russian to Chinese of such stereotypical associations as \texttrait{egoistic, threatening, repellent}, and \texttrait{cold}, while in the human data in Hindi, this group is perceived as \texttrait{warm}. 

Another example is \textgroup{immigrants}. From Russian and English languages, traits such as \texttrait{threatening, repellent, dishonest, egoistic}, and \texttrait{unconfident} leak to Chinese and Hindi. Based on human data, we found that people surveyed in Chinese view this group quite favorably since the majority of immigrants to China were highly qualified professionals \citep{Pieke_2012}. Contrarily, in Russia, immigrants are mostly coming from poorer neighboring countries and are negatively stereotyped in society, while in the U.S., immigrants are diverse and could be both marginalized or privileged.
 
Moreover, there is a notable leakage from English to Chinese and Hindi for \textgroup{Black people} for traits \texttrait{dominated} and \texttrait{poor}. This aligns with known stereotypes about African Americans and Africans in U.S. society \citep{miller, galster,beresford}. 

\subsubsection{Non-polar Leakage}

There are also non-polar leakages, which are neither positive nor negative. From Hindi to English and Russian, we see the strengthening of \texttrait{religious} for various groups such as \textgroup{women}, \textgroup{disabled people}, \textgroup{Black people}, and \textgroup{Asian people}. It has been shown that there are more than $70.00 \%$ believers of the total population in India as of $2011$\citep{evans2021key}.

\subsubsection{Non-shared Groups Leakage}

In the case of non-shared groups, we expected uni-directional transferring of the groups' perceptions from the language of origin to other languages. Our findings confirm this hypothesis. For example, the group \textgroup{VDV soldiers} is a widely known military unit in Russia. There are strong stereotypes in Russian society about this group, but the group is mostly unknown to Americans. Out of the $34$ survey English survey respondents who passed the quality tests, no one chose this group as a familiar one. Stereotypes of this group leak from Russian to English, strengthening traits such as \texttrait{confident, traditional, competitive}, and \texttrait{threatening}. Another example is the \textgroup{Hui people}, a group widely unknown to Russian and Hindi society: out of $76$ respondents for both surveys, no one chose this group as the familiar one. This social group is a minority in China and is composed of Chinese-speaking followers of Islam. Originally, \textgroup{Hui people} were marginalized in China and viewed as more traditional, religious, and conservative \citep{hillman2004rise, hong}. Accordingly, we observed the leakage of such traits as \texttrait{irrational, traditional, threatening, repellent, religious, and egoistic}. All groups specific to the Hindi language --- \textgroup{Gujarati, Brahmin}, and \textgroup{Shudra people} --- have certain traits leaking to the English and Russian languages. For example, high caste groups (\textgroup{Gujarati} and \textgroup{Brahmin people}) strengthen such positive traits as \texttrait{wealthy, likable, sincere, powerful, high status, competitive}, and \texttrait{confident}. In addition, \textgroup{Brahmin people} become more associated in GPT-3.5 with traits \texttrait{poor, low status, powerless, traditional, religious}, and \texttrait{dominated}. This leakage corresponds to the perception of these groups in Indian society and by our survey respondents \citep{Witzel1993TowardAH, milner}.

\subsection{Discussion}
The amplification of negative stereotypes is certainly a cause for concern. These stereotypes, often deeply ingrained in societal narratives, can perpetuate discrimination and prejudice. Conversely, while positive stereotypes might seem harmless or even beneficial at first glance, they can also be problematic. In some contexts, positive stereotypes may serve to counterbalance negative ones, creating what is known as an anti-stereotype effect. This can be useful in mitigating some of the harms caused by negative stereotypes.

However, positive stereotypes, such as the notion that \textgroup{Asian people} are \texttrait{wealthy} or \textgroup{housewives} are \texttrait{warm}, can also lead to unrealistic expectations and pressures. For instance, not all Asian people are wealthy, and assuming so can ignore the diverse economic realities faced by individuals within this broad demographic. Similarly, the stereotype that housewives are inherently warm can enforce restrictive roles based on gender.

The leakage of stereotypes is particularly troubling for certain applications, including education and creative content generation. These fields heavily influence public perception and personal development, making the integrity of the content they deliver crucial. Systems built for these applications with MLLMs must be particularly cautious of the stereotype leakage effect. Developers need to implement strategies that actively mitigate the harmful leakage effects.

\section{Conclusion }

Multilingual large language models have the potential to spread stereotypes beyond the societal context they emerge from, whether by generating new stereotypes, amplifying existing ones, or reinforcing prevailing social perceptions from dominant cultures. In our study, we demonstrate that this concern is indeed valid. To do so, we establish a framework for measuring the leakage of stereotypical associations in multilingual large language models across languages. Overall, we find that the stereotype leakage occurs bidirectionally meaning that when one language transmits stereotypes to others, it likely receives some stereotypes from other languages as well. We also observe the most stereotype leakage effect within the GPT-3.5 model. 

Within the GPT-3.5 model, we observe the strengthening of positive, negative, and non-polar associations in the model. In addition, our study underscores the role of ``native'' languages in framing social groups unknown to other linguistic communities. Such leakage of stereotypes amplifies the complexity of societal perceptions by introducing a complex interconnected bias from different languages and cultures. In the context of shared groups, stereotype leakage may manifest as the manifestation of stereotypes that were not previously present within the cultural setting of a particular group. For non-shared groups, stereotype leakage can extend the reach of existing stereotypes from the source culture to other cultural contexts.

To our knowledge, we are the first to introduce the concept of stereotype leakage across languages in multilingual LLMs. We propose a framework for quantifying this leakage in multilingual models, which can be easily applied to unstudied social groups. We show that multilingual large language models could facilitate the transmission of biases across different cultures and languages. We demonstrate the existence of stereotype leakage within MLLMs, which are trained on diverse linguistic datasets. As multilingual models begin to play an increasingly influential role in AI applications and across societies, understanding their potential vulnerabilities and the level of bias propagation across linguistic boundaries becomes important. As a result, we lay the groundwork for advancing both the theoretical comprehension of multilingual models and the practical implementation of bias mitigation in AI systems.

\section*{Limitations and Ethical Considerations}
Our study has several limitations. 
First, we are limited in our ability to run a causal analysis because none of the studied languages can be easily removed from the training data to see their genuine impact on stereotypical associations in other languages. Retraining GPT-3.5, for instance, is not a feasible option. Thus, we use the BERT monolingual model as a proxy for each language.

In addition, stereotype traits were selected based on the ABC model, which was developed and tested using U.S. and German stereotypes. Though we translated our surveys into all four languages, the stereotype traits may better reflect Anglocentric stereotypes \citep{talat2022you} than others.

Furthermore, the human stereotypes we collected may already reflect the influence of social stereotype transmission. For instance, in our study, we surveyed crowd workers about their consumption of U.S. social media. We found that, on average, $39\%$ of respondents from Russia, China, and India engage with U.S. social platforms. Such American cultural dominance could affect the human stereotypes collected in these three languages.

Lastly, while we indirectly consider culture through survey results on associations, we do not measure or account for culture comprehensively. Our English language survey results only apply to the U.S., Russian to Russia, Chinese to China, and Hindi to India.






\begin{acks}
This work is supported by the National Science Foundation under Grant No. 2131508, as well as Grant Nos. 2229885, 2140987, and 2127309 awarded to the Computing Research Association for the CIFellows Project. We express our gratitude to Chenglei Si for prompting suggestions, Navita Goyal for her assistance with translations, and the members of the Clip Lab at the University of Maryland, along with our friends, for their contributions to the pilot surveys.
\end{acks}


\bibliographystyle{ACM-Reference-Format}
\bibliography{sample-base,stereo}

\clearpage
\appendix
\newpage
\section{Human Study}\label{sec:survey}
In this Section, we present details about the survey design, annotations quality control, and participants' demographics.
\subsection{Survey Design}\label{sec:design}
\begin{figure}[ht]
\centering
\includegraphics[width=1\columnwidth]{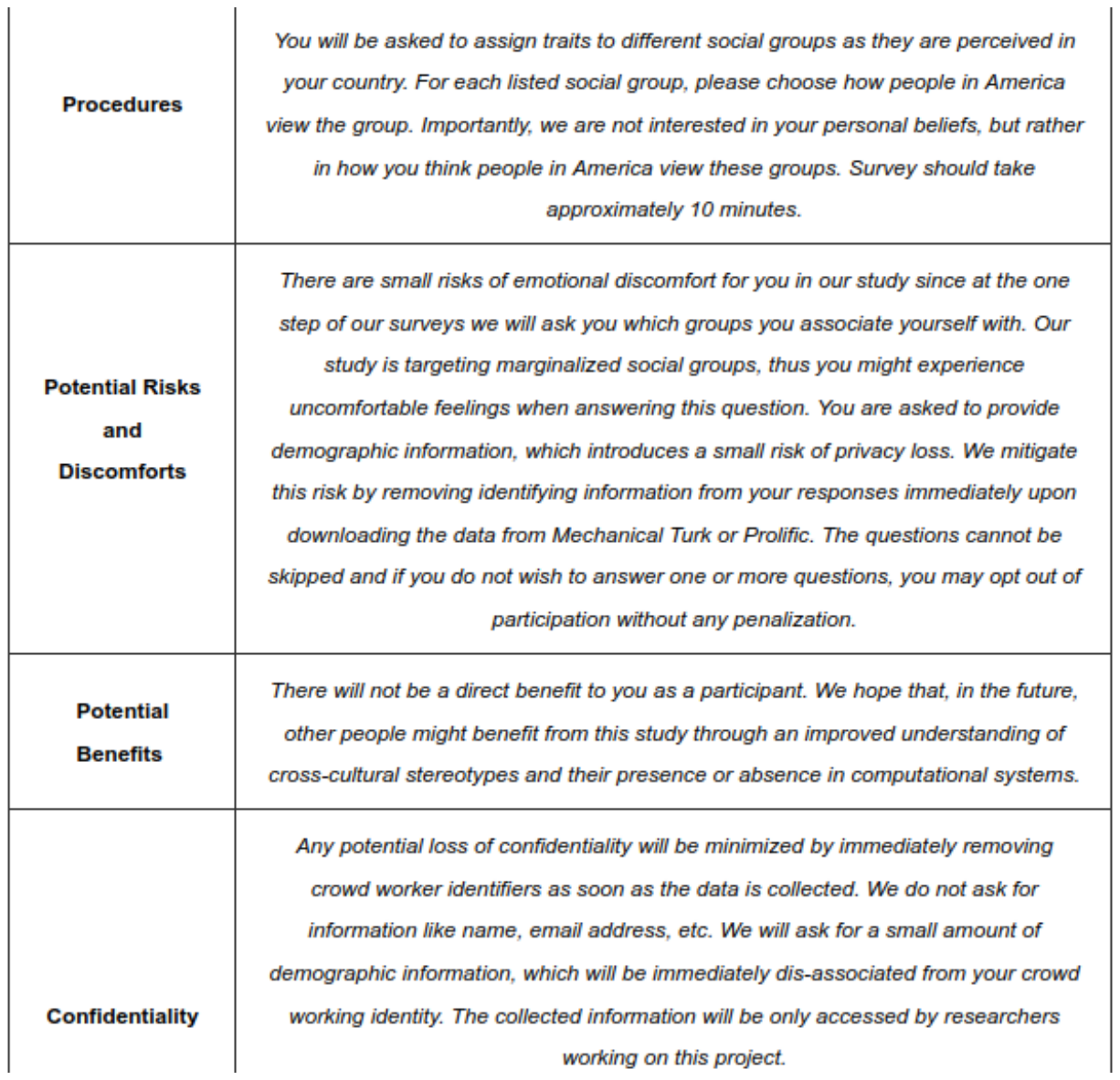}
\caption{Selected points of the consent form highlighting study format, confidentiality, and potential risks.}
\label{fig:consent}
\end{figure}

Participants first reviewed a consent form, which outlined the purpose of the study, data usage, and confidentiality. Only after agreeing to participate, they proceed to the survey instructions. The consent form is shown in Figure \ref{fig:consent}

\begin{figure}[t]
\centering
\includegraphics[trim=0cm 0.5cm 0cm 0cm, clip, width=1\columnwidth]{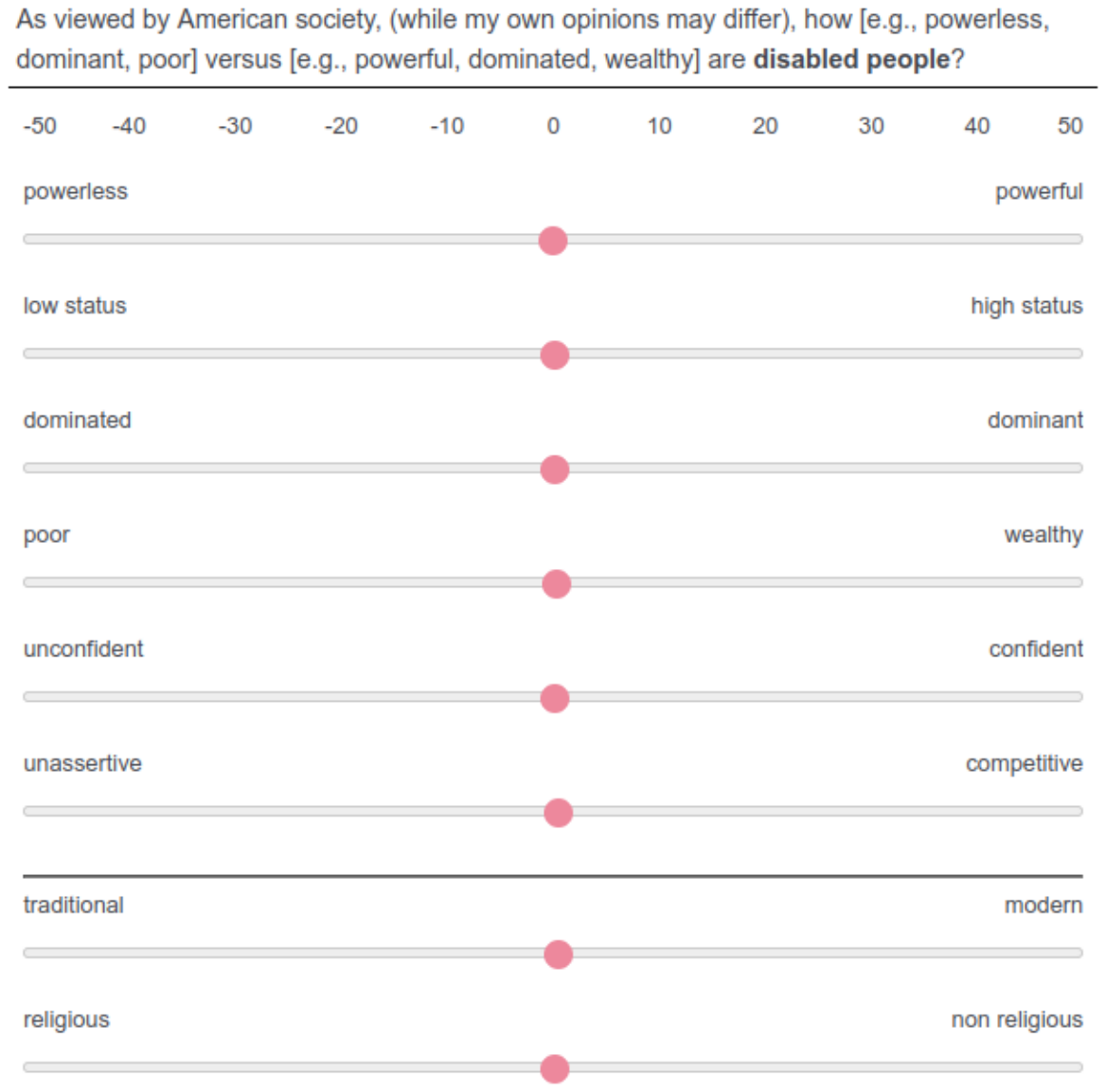}
\caption{Example of the survey.}
\label{fig:survey2}
\end{figure}

For each social group, participants read the following prompt in their respective language: ``As viewed by American/Russian/Chinese/Indian society, (while my own opinions may differ), how \texttt{[e.g., powerless, dominant, poor]} versus \texttt{[e.g., powerful, dominated, wealthy]} are \texttt{<group>}?'' They then rated each group on a slider scale ranging from -50 to 50, where the two poles of the scale represented opposite traits (e.g. \texttrait{powerless} and \texttrait{powerful}). Each social group appeared on a separate page, and participants were unable to return to previous pages, helping to minimize response bias. An example of the task is presented in Figure \ref{fig:survey2}

\begin{figure}[t]
\centering
\includegraphics[trim=2cm 9cm 2cm 1cm, clip, width=1\columnwidth]{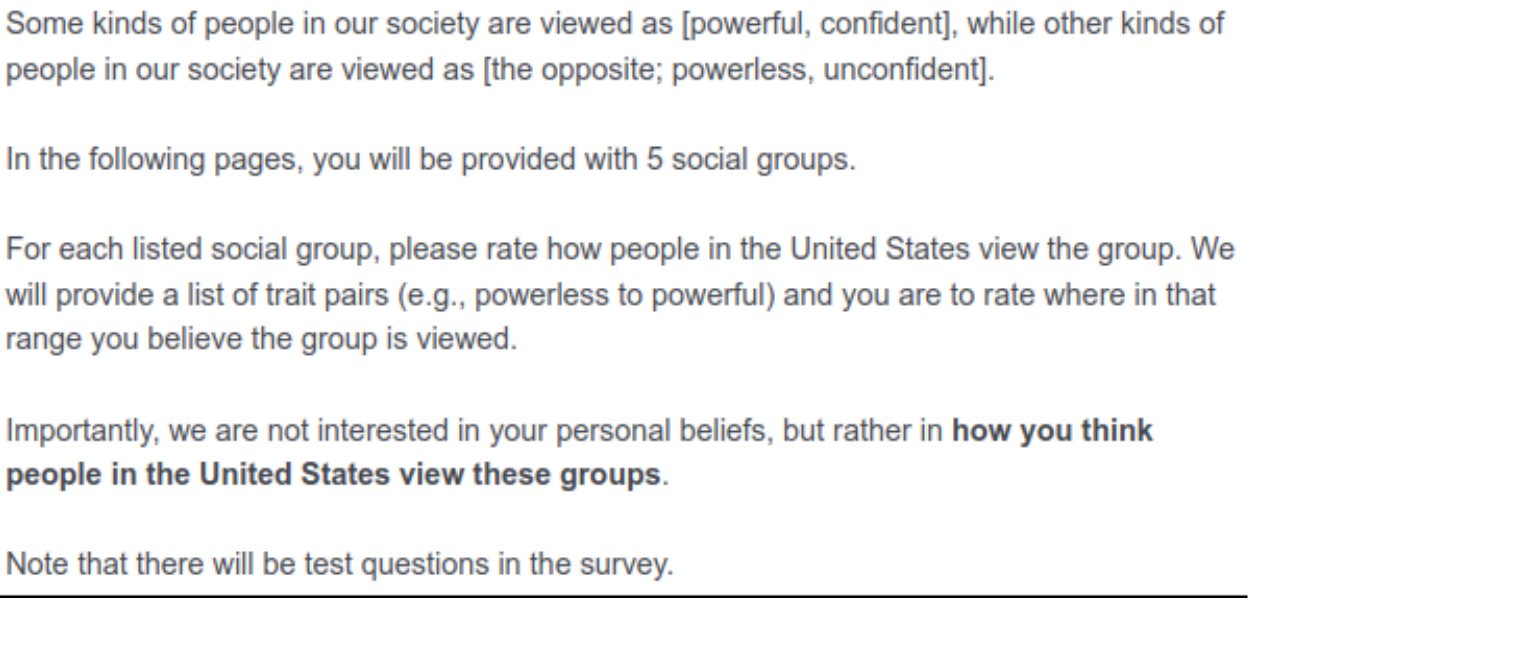}
\caption{Instructions before crowd workers view the task itself.}
\label{fig:survey1}
\end{figure}

To reduce social desirability bias, the instructions clearly emphasized: ``We are not interested in your personal beliefs, but rather in how you think people in the United States/Russia/China/India view these groups.'' The exact formulation is presented in Figure \ref{fig:survey1}.

Participants were paid \$2.00 to rate five social groups on 16 pairs of traits, which took an average of 10 minutes to complete, translating to a compensation rate of \$12.00 per hour. 


\subsection{Quality Assurance}\label{sec:quality}

Collecting high-quality data for subjective tasks presents significant challenges, particularly due to the absence of objective ground truth. To mitigate these challenges, we implemented rigorous quality control procedures to ensure reliability and consistency across annotations.

The survey was administered through the Prolific platform, and only participants with an approval rate exceeding 90\% were eligible to participate. This threshold was selected to balance data quality with participant availability, as it is generally considered high for Prolific, increasing the likelihood of obtaining reliable data.

In addition to the platform's approval rate, we implemented three test questions throughout the survey to assess attentiveness and comprehension:
\begin{itemize}
     \item After the first group, participants must name the group they just scored.
\item After the second, participants must list one trait they just marked high and one marked low.
\item The fifth (final) group is a repetition of one of the four groups they previously scored.
\end{itemize}
We exclude annotators who answered the first two questions incorrectly. We then measured their intra-annotator (self) agreement by comparing the consistency of their responses, and any annotation with less than 80\% self-agreement was discarded.
These measures helped ensure data quality, though all participants were compensated regardless of their performance in the quality tests.

We collected at least five valid annotations per group that met our quality thresholds. Of the 286 participants, 151 passed the quality checks. Specifically, 34 participants passed for the English-language survey, 36 for Russian, 41 for Chinese, and 40 for Hindi. The fact that nearly half of the participants failed to meet the quality criteria underscores the necessity of these controls in subjective data collection. 

\subsection{Participant Demographics}\label{sec:demo}

We collected demographic information from participants, including gender, age, education level, and, for non-English speakers, their frequency of reading American social media. Participants were free to skip any question they preferred not to answer.

Across all languages, the gender distribution revealed a near balance: 49\% identified as male, 45\% as female, and 5\% as non-binary, transgender, or gender fluid, with a few opting not to disclose. When we examined educational backgrounds, participants from non-English-speaking countries showed similar trends: 36\% held a bachelor’s degree, 32\% had a master’s degree, and 7\% had earned a Ph.D. The remaining respondents either had lower educational qualifications or chose not to answer. English-speaking participants stood out, with no respondents holding a Ph.D., a lower percentage with master’s degrees (29\%), and a larger proportion (35\%) being high school graduates.

Among the English-speaking survey group, the largest proportion of respondents hailed from Texas (15\%), followed by California and New York (each contributing 9\%). The remaining participants were dispersed across 25 states, with no significant regional concentration outside these key areas.

As we looked at the age distribution, it was clear that younger people dominated the study, with 42\% aged between 18 and 30 and 33\% falling in the 31 to 40 range. The remainder were above 40, with the youngest participant being 18 and the oldest, a more experienced 72. 

One notable demographic trend emerged in media consumption habits. Russian-speaking participants were the most frequent consumers of American media, with 44\% stating they read it regularly. In contrast, 35\% of Hindi-speaking participants and 28\% of Chinese-speaking participants reported similar habits. Across all groups, about 39\% said they occasionally consumed American media, while only 5\% never did. These patterns suggest that Russian participants may be more exposed to or interested in global perspectives, particularly through American social media.

Crucially, all approved participants confirmed fluency in the language of their respective surveys. This ensures that any differences in responses were not influenced by language proficiency but more likely reflect deeper cultural or regional perspectives.

\section{Model Stereotypical Association Measurement}
\subsection{GPT Model Generation Failures on Marginalized Groups}\label{sec:failure}
We observe that for certain groups like \textgroup{feminist} and \textgroup{Muslim person} in Chinese, the model often disregards the prompt and simply outputs the group name. Moreover, in some cases, the model alters the trait specified in the prompt. For example, it changes \texttrait{dominating} to \texttrait{dominated} for \textgroup{disabled person} in English or \texttrait{poor} to \texttrait{wealthy} for \textgroup{migrant worker} in Russian. Additionally, the model may overlook the traits provided in the prompt and generate stereotypical traits instead. For instance, in Russian, it generates \texttt{rape} and \texttt{patriot} for \textgroup{Puerto Rican} or \texttt{cowboy} for \textgroup{Texan}.

We also count the number of system generations that did not match the instruction requirements for each social group. For example, in Chinese, we observed 108 generations for the group ``feminist'' that did not match the instruction requirement out of a total of 2880 generations. In comparison, there were 20 non-matching generations for ``women.'' However, these figures represent only an upper bound of system failures, as various reasons, such as generating synonyms, could cause mismatches. As stated in the paper, we leave the in-depth analysis for future work.

\end{document}